\documentclass[accepted]{uai_arxiv} % for arxiv
\usepackage[american]{babel}
\usepackage{natbib} % has a nice set of citation styles and commands
\usepackage{mathtools} % amsmath with fixes and additions
\usepackage{booktabs} % commands to create good-looking tables
\usepackage{tikz} % nice language for creating drawings and diagrams
\DeclareMathAlphabet{\mathcal}{OMS}{cmsy}{m}{n}
\usepackage{framed,multirow}
\usepackage{amssymb}
\usepackage{hyperref}

\DeclareMathOperator*{\argmax}{arg\,max}

\newcommand{\vecs}[1]{\mbox{\boldmath $#1$}}
\usepackage[]{xcolor}

\usepackage[caption=false,font=footnotesize]{subfig}

\title{Similarity Measure for Sparse Time Course Data Based on Gaussian Processes}

\author[1,2]{\href{mailto:Zijing Liu <zijing.liu@imperial.ac.uk>?Subject=Your UAI 2021 paper}{Zijing Liu}{}} % Lead author
\author[2]{Mauricio Barahona}
\affil[1]{%
    Department of Brain Sciences\\
    Imperial College London\\
    London, UK
}
\affil[2]{%
    Department of Mathematics\\
    Imperial College London\\
    London, UK
}

\begin{document}
\maketitle

\begin{abstract}
We propose a similarity measure for sparsely sampled time course data in the form of a log-likelihood ratio of Gaussian processes (GP). The proposed GP similarity is similar to a Bayes factor and provides enhanced robustness to noise in sparse time series, such as those found in various biological settings, e.g., gene transcriptomics. We show that the GP measure is equivalent to the Euclidean distance when the noise variance in the GP is negligible compared to the noise variance of the signal. Our numerical experiments on both synthetic and real data show improved performance of the GP similarity when used in conjunction with two distance-based clustering methods.
\end{abstract}

\section{Introduction}

Time course data are used widely for the empirical study of dynamical processes in many areas of research in the natural and social sciences~\citep{keogh2003need}.
Traditionally, much research has been devoted to the characterisation of time series in relation to the originating dynamical process from different viewpoints, from the deterministic to the stochastic~\citep{brillinger1981time,barahona1996,chatfield2003analysis}. 

More recently, time series have also been considered from the perspective of data science. One of the key questions in many applications is to assess the (dis-)similarity between time courses, with a view to perform time series classification or clustering~\citep{liao2005clustering,son2008modified,gorecki2014using,Peach2019,fulcher2014highly}.
%\MB{Add Nick Jones paper on htsa}
A simple way to deal with finite, discretely-sampled time courses is to treat them as vectors~\citep{yao2005functional,hedeker2006longitudinal}, i.e., a time series sampled at $t$ time points $\{y_{i}(t_k):=y_{ik}\}_{k=1}^t$ is described by a $t$-dimensional vector $\mathbf{y}_i$ with coordinates $y_{ik}$. 
%and $\mathbf{y}_j$ sampled at $t$ time points $y_{ik}, y_{jk}$.
A simple dissimilarity measure between two time series $\mathbf{y}_i$ and $\mathbf{y}_j$ is then given by the Euclidean ($\ell^2$) distance:
\begin{equation}
\label{eq:Euclidean}
    d_{2}(\mathbf{y}_i,\mathbf{y}_j) = \sqrt{\sum_{k=1}^t \left(y_{ik}-y_{jk}\right)^2} = \Vert \mathbf{y}_i-\mathbf{y}_j \Vert.
\end{equation}
%where  
%are the $k$-th component of $\mathbf{y}_i$ and $\mathbf{y}_j$, respectively. 
Alghouth the Euclidean distance is widely used due to its simplicity, 
%but it does not capture global trends in the data. 
in some applications, one may be more interested in the trend of how the data changes across time rather than the absolute differences. To capture this, a frequently used dissimilarity measure is based on Pearson's correlation coefficient, $r_{\mathbf{y}_i \mathbf{y}_j}$:
\begin{align*}
 & d_\text{corr}(\mathbf{y}_i,\mathbf{y}_j) = 
 1 - r_{\mathbf{y}_i \mathbf{y}_j}  \\
 & \qquad =  1 - \frac{ \sum_{k=1}^t (y_{ik}-\overline{y}_i)(y_{jk}-\overline{y}_i)}{\sqrt{\sum_{k=1}^t (y_{ik}-\overline{y}_i)^2}\sqrt{\sum_{k=1}^t (y_{jk}-\overline{y}_j)^2}} ,
\end{align*}
where $\overline{y}_i$ and $\overline{y}_j$ are the means of each time series.
Note that these two common measures are point-to-point matching, and insensitive to the re-ordering of the time points and the spacing between the sampling times.
%Although time course data can be equally spaced across time, it is ubiquitous that the time interval is variable. Thus, the point-wise distance
Hence these measures cannot capture information associated with the time indices, which can be important in applications.  

To remedy the limitations of point-wise measures, alternative measures of dissimilarity in time series have been proposed, including Dynamic Time Warping (DTW)~\citep{keogh2005exact} and the Edit Distance on Real Sequences (EDR)~\citep{chen2005robust}. These methods can cope with uneven sampling and use information from the time indices, yet they can be algorithmically complex and are not well suited for applications with a large number of short, sparsely sampled time courses. Examples of this type of data are common in home price, marketing or e-commerce data in economics and finance~\citep{fan2011sparse}, longitudinal electronic healthcare records in healthcare~\citep{perotte2013temporal}, genomics and proteomics data in life science~\citep{ndukum2011statistical,kayano2016gene}, and functional magnetic resonance imaging~\citep{smith2012future}.
For instance, in cellular biology, `omics' experiments 
%use different techniques (e.g., RNAseq, mass spectrometry, etc) to 
measure the expression level of large numbers of genes, proteins or metabolites in cells over time. Such datasets contain tens of thousands of time courses (i.e., the number of genes or proteins)
%$10^3$ to $10^4$) 
but the length of each time course is very short (5-15 time points) due to the high cost of experiments. Furthermore, the time samplings are usually uneven since experiments are designed to capture trends in cellular evolution and responses to stimuli. These constraints are typical of many biological experimental settings.

In this paper, we introduce a similarity measure for time course data based on Gaussian processes (GPs)~\citep{rasmussen2006gaussian}, which is applicable to sparse, inhomogeneously sampled, high-dimensional datasets. %
To retain information from the sampling times in the data, we model the time courses as continuous functions using GPs, and define a similarity measure in the form of a log-likelihood ratio between GP models. 
The GP similarity is computationally simple and suitable for high-dimensional datasets with a large number of short time courses. We also show that the GP similarity measure is equivalent to the Euclidean distance when the noise variance in the GP model is negligible compared to the signal variance. We apply the GP similarity measure as the basis for distance-based clustering methods in both synthetic and real time course data, and show improved robustness to measurement noise and to sampling inhomogeneity.

\section{Related work}
As a non-parametric model, GP is a flexible and efficient tool for time-dependent data modelling. 
Using the fact that a GP defines a reproducing kernel Hilbert space (RKHS), \cite{lu2008reproducing} proposed a RKHS-based distance for time series defined as the Bregman divergence between the two posterior GPs. This distance has a closed form: it is the squared norm of the posterior mean functions in the RKHS induced by the GP. However, the Bregman divergence does not reflect the uncertainty of the data, since it only depends on the posterior mean function. Hence the Bregman divergence can perform poorly in the presence of  noise in the data, as we show below.

GPs have been previously applied to time series of gene expression to detect differentially expressed genes~\citep{stegle2010robust,kalaitzis2011simple} and to infer the dynamics of transcriptional regulation~\citep{lawrence2007modelling,gao2008gaussian}. In~\cite{kalaitzis2011simple}, the fitted GP model for each gene is compared to a noise model in order to rank the time courses and find differentially expressed genes. Here, we use the construction of GPs differently, and show that the likelihood ratio between two GP models provides a robust similarity measure for time courses. In the next section, we will introduce the GP model for time course data.

%\MB{I know that Gretton has also done work on functional analysis using kernels. Worth mentioning any of his papers??}

\section{Gaussian process model for time course data}

A Gaussian process is a collection of random variables over the index set $\mathcal{X}$ such that any finite collection of the random variables follows a multivariate normal distribution~\citep{rasmussen2006gaussian}.
Therefore, a Gaussian process, denoted $\mathcal{GP}(m(x),k(x,x'))$, is characterised by the mean function $m(x)$ and the covariance function $k(x,x')$, where $x,x' \in \mathcal{X}$.
Here we will consider time-dependent variables; hence the index set is the positive real line describing time:
$\mathcal{X}=\mathbb{R}^+$.

We will model the underlying true signal as a Gaussian process over time:
\begin{equation}
\label{eq:GP_eqn}
   f(x) \sim \mathcal{GP}\left(m(x),k(x,x')\right).
\end{equation}
For simplicity, we take the mean function to be zero ($m(x)=0$) and we use the squared exponential covariance function
\begin{equation}
\label{eq:Gaussian_covariance}
    k(x,x') = \sigma_f^2 \, \mathrm{exp} \left( -\frac{\Vert x-x' \Vert ^2 }{2 \ell^2} \right) =: \sigma_f^2 \, G(x,x')
\end{equation}
where $\ell$ is a characteristic length-scale, $\sigma_f^2$ is the signal variance and we use $G(x,x')$ to denote the Gaussian kernel.
The observations of the time-dependent variable are then noisy samples of the GP:
\begin{equation}
\label{eq:GP_eq}
    y=f(x)+\varepsilon,
\end{equation}
where the additive noise $\varepsilon$ is Gaussian with zero mean and variance $\sigma_n^2$.

Let us consider a time-dependent variable $y$ given by~\eqref{eq:GP_eq} sampled at $t$ time points $X = [x_1,\dots,x_t]^T$ and let us compile the observations into a $t$-dimensional vector $\mathbf{y}=[y_1, \ldots, y_t]^T$. 
Under our assumptions, the covariance function of the noisy observations $y$ is given by:
\begin{align}
\label{eq:covariance}
    k_y (x_p,x_q)
 %   &= \sigma_f^2 \, \mathrm{exp}\left( -\frac{\Vert x_p-x_q \Vert ^2}{2 \ell^2} \right) + \sigma_n^2 \, \delta_{pq} \\
  & = \sigma_f^2 \, G(x_p,x_q) + \sigma_n^2 \, \delta_{pq}, 
 %\\
%  & = k(x_p,x_q) + \sigma_n^2 \, \delta_{pq} \nonumber
\end{align}
where $\delta_{pq}$ is the Kronecker delta. Equivalently, the $t \times t$ covariance matrix of the observations $\mathbf{y}$ is:
\begin{equation}
\label{eq:covariance_matrix}
    K_y = \sigma_f^2 \, G +  \sigma_n^2 \, I = K +  \sigma_n^2 \, I, 
\end{equation}
where $G$ is the Gaussian kernel matrix with elements $G_{pq} = G(x_p,x_q)$, $I$ is the identity matrix of dimension $t$, and $K$ is the covariance matrix for the noiseless samples with elements $K_{pq} = k(x_p,x_q)$.
%The covariance matrix of the observations is $\mathcal{K}=K +  \sigma_n^2 I$ if $I$ is an identity matrix.

The three hyperparameters of the Gaussian process are therefore $\vecs{\theta} = (\ell,\sigma_f,\sigma_n)$, and can be learnt from the data $(X,\mathbf{y})$ by maximising
\begin{equation}
    \vecs{\theta}^* = \argmax_{\vecs{\theta}} 
   \log p(\mathbf{y}|X,\vecs{\theta}),
\end{equation}
where, in this case, the log-marginal likelihood has the explicit form:
\begin{align}
    %  &\log p(\mathbf{y}|X,\vecs{\theta}) \\ &=-\frac{1}{2} \mathbf{y}^T (K +  \sigma_n^2 I)^{-1} \mathbf{y} - \frac{1}{2} \log \vert K +  \sigma_n^2 I \vert - \frac{t}{2} \log 2\pi.
\log p(\mathbf{y}|X,\vecs{\theta}) =-\frac{1}{2} 
\left(\mathbf{y}^T K_y^{-1} \mathbf{y} + \log \det K_y  + t \, \log 2\pi \right),
\label{eq:Gaussian_likelihood}
\end{align}
with $\det K_y$ denoting the determinant of $K_y$.
This expression can be maximised using gradient-based methods~\citep{rasmussen2006gaussian}. 

\section{Similarity measure for time course data based on Gaussian process}

Let us consider a sparse time course dataset consisting $N$ short time courses sampled at $t$ time points: $\mathbf{y}_i \in \mathbb{R}^t, \, i=1,\ldots,N$. The dataset is referred to as sparse, due to the fact that the number of time courses is much larger than the length of each time course ($N \gg t$).
Although the GP model does not require all the time series to be measured synchronously, for simplicity, we first introduce the similarity measure for the case where all time courses are sampled at the same time points $X = [x_1,\dots,x_t]^T$. We discuss the asynchronous case later.

Each of the time courses $\mathbf{y}_i$ is assumed to correspond to a noisy observation of a Gaussian process~\eqref{eq:GP_eq} with the same hyperparameters~$\boldsymbol{\theta}$, 
%\MB{Is this correct??  All the $N$ series are described by the same GP??  What about the technical replicates that we also have?  How do they appear?}
%
which can be inferred by maximising the sum of the log marginal likelihoods of the time courses~\citep{rasmussen2006gaussian}: 
%If there are $n$ time courses and all the time courses have the same time points $X$, the hyperparameters can be learned by 
\begin{equation}
\label{eq:inference_highdim}
    \vecs{\theta}^* = \argmax_{\vecs{\theta}} \sum_{i=1}^N \log p(\mathbf{y_i}|X,\vecs{\theta}).
\end{equation}
By inferring the hyperparameters $\boldsymbol{\theta}^*$, we obtain a non-parametric probabilistic model for the observed time courses, which we can use to define a GP-based similarity measure, as follows.

\subsection{Likelihood ratio as a simple GP similarity measure}
Using the fact that the GP is a distribution of continuous functions over time, a similarity measure between two observed time courses $\mathbf{y}_i$ and $\mathbf{y}_j$ can be obtained by comparing two different possibilities as to how $\mathbf{y}_i$ and $\mathbf{y}_j$ could have been generated.
%which is a distribution of continuous functions. 

The first possibility is that the two time samples $\mathbf{y}_i$ and $\mathbf{y}_j$ are observations from \textit{two different} functions sampled from the GP.
In this case, the joint likelihood of $\mathbf{y}_i$ and $\mathbf{y}_j$ is just the product of the likelihoods of two time courses: 
\begin{equation}
    p_\text{diff}(\mathbf{y}_i,\mathbf{y}_j|\vecs{\theta}^*)= p(\mathbf{y}_i|X,\vecs{\theta}^*) \, p(\mathbf{y}_j|X,\vecs{\theta}^*).
    \label{eq:lik_independent}
\end{equation}
Using~\eqref{eq:Gaussian_likelihood}, it is easy to see that the log-likelihood can be rewritten as:
\begin{align}
 %   \begin{split}
        & \log p_\text{diff}(\mathbf{y}_i,\mathbf{y}_j|\vecs{\theta}^*) =
        \log p(\mathbf{y}_i|X,\vecs{\theta}^*) + \log p(\mathbf{y}_j|X,\vecs{\theta}^*) \nonumber \\
       & \hskip 30pt = -\frac{1}{2}
       \begin{bmatrix}
    \mathbf{y}_i^T & \mathbf{y}_j^T
    \end{bmatrix} 
    \begin{bmatrix}
      K_y & 0  \nonumber \\
    0 & K_y
    \end{bmatrix}^{-1}
    \begin{bmatrix}
    \mathbf{y}_i \\ 
    \mathbf{y}_j
    \end{bmatrix} \\
    & \hskip 40pt -\frac{1}{2} \log \det
    \begin{bmatrix}
    K_y & 0 \\
    0 & K_y 
    \end{bmatrix}
    -  t \, \log 2\pi.
 %   \end{split}
 \label{eq:likelihood_diff}
\end{align}

The second possibility is that $\mathbf{y}_i$ and $\mathbf{y}_j$ are observations from \textit{the same} function sampled from the GP. 
In this case, the joint likelihood of $\mathbf{y}_i$ and $\mathbf{y}_j$ can be computed by considering the two time courses $\{X,\mathbf{y}_i \}$ and $\{X,\mathbf{y}_j \}$ to be replicate samples of one function
%, i.e., $\left\{\begin{bmatrix}
%     X\\ 
%     X
%     \end{bmatrix},
%     %[\mathbf{y}_i ; \mathbf{y}_j],
%     \begin{bmatrix}
%     \mathbf{y}_i \\ 
%     \mathbf{y}_j
%     \end{bmatrix} \right\}$
:  
%\MB{Concatenating means that the last time point gets linked to the first time point of the second series. Is this what we do? If so, there are edge effects then by matching end-beginning of the series?}
\begin{equation}
    p_\text{same}(\mathbf{y}_i,\mathbf{y}_j|\vecs{\theta}^*) = p \left(
    \begin{bmatrix}
    \mathbf{y}_i \\ 
    \mathbf{y}_j
    \end{bmatrix} \bigg\vert
    \begin{bmatrix}
    X\\ 
    X
    \end{bmatrix},\vecs{\theta}^*\right),
    \label{eq:lik_combine}
\end{equation}
and the log-likelihood is then given by:
\begin{equation}
    \label{eq:likelihood_same}
\begin{split}
    &\log p_\text{same}(\mathbf{y}_i,\mathbf{y}_j|\vecs{\theta}^*) = \log p\left(
    \begin{bmatrix}
    \mathbf{y}_i \\ 
    \mathbf{y}_j
    \end{bmatrix}\bigg\vert 
    \begin{bmatrix}
    X \\ 
    X
    \end{bmatrix},\vecs{\theta}^*\right) \\ 
    & \hskip 30pt = - \frac{1}{2} 
    \begin{bmatrix}
    \mathbf{y}_i^T & \mathbf{y}_j^T
    \end{bmatrix} 
    \begin{bmatrix}
      K_y & K  \\
    K & K_y
    \end{bmatrix}^{-1}
    \begin{bmatrix}
    \mathbf{y}_i \\ 
    \mathbf{y}_j
    \end{bmatrix} \\
    & \hskip 40pt -\frac{1}{2} \log \det
    \begin{bmatrix}
    K_y & K\\
    K & K_y 
    \end{bmatrix}
    -  t \, \log 2\pi,
\end{split}
\end{equation}
where we have used the fact that the additive noise $\varepsilon$ in~\eqref{eq:GP_eq} is uncorrelated between the two time courses.

\begin{figure}[!th]
\centering
\subfloat[Two time courses with similar patterns and the corresponding GP.]
{\includegraphics[width=0.9\linewidth]{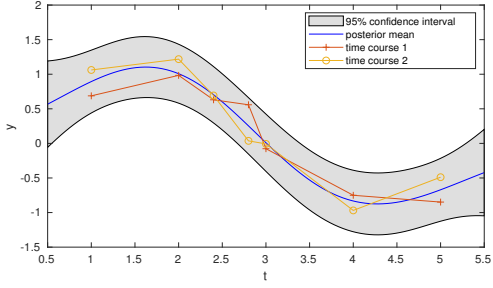}%
\label{fig:bf1}}
\hfil
\subfloat[Two time courses with different patterns and the corresponding GP.]
{\includegraphics[width=0.9\linewidth]{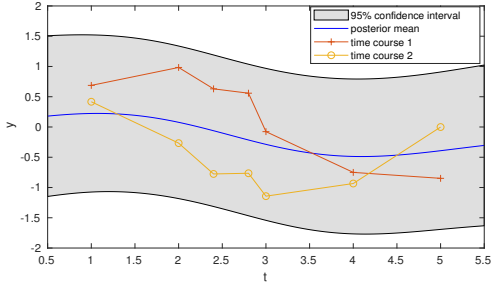}%
\label{fig:bf2}}
\caption{Two examples with two time courses that have: (a) similar and (b) dissimilar profiles. In each case, we show the time courses and the mean and confidence interval of the Gaussian process~\eqref{eq:lik_combine} obtained according to the likelihood in~\eqref{eq:lik_combine}.
%\MBB{Why not show also what happens when you fit them independently with two GPs, as in \eqref{eq:lik_independent}??}
}
\label{fig:bf12}
\end{figure}

The likelihood~\eqref{eq:lik_combine} will be high if the two time courses are similar to each other (as in Fig.~\ref{fig:bf1}), and will be small if the two time courses have different profiles (as in Fig.~\ref{fig:bf2}). Hence for time courses with different profiles, the likelihood~\eqref{eq:lik_independent} explains better the data. 
The log of the ratio of the two likelihoods~\eqref{eq:lik_combine}--\eqref{eq:lik_independent} (i.e., the difference between the log-likelihoods) is thus an indicator of the level of similarity between two time courses. This leads to our definition of the \textit{GP similarity measure} between $\mathbf{y}_i$ and $\mathbf{y}_j$ as:
\begin{align}
 %\begin{split}
& s(\mathbf{y}_i,\mathbf{y}_j) = 
 \log \frac{ p_\text{same}(\mathbf{y}_i,\mathbf{y}_j|\vecs{\theta}^*)}
{p_\text{diff}(\mathbf{y}_i,\mathbf{y}_j|\vecs{\theta}^*)} 
%=\log \frac{p([\mathbf{y}_i;\mathbf{y}_j]|[X;X],\vecs{\theta})}{p(\mathbf{y}_i|X,\vecs{\theta}) \ p(\mathbf{y}_j|X,\vecs{\theta})} 
\label{eq:bayesfactor1}
\end{align}
and it follows that:
\begin{align}
s(\mathbf{y}_i,&\mathbf{y}_j) =\log p_\text{same}(\mathbf{y}_i,\mathbf{y}_j|\vecs{\theta}^*) -
\log p_\text{diff}(\mathbf{y}_i,\mathbf{y}_j|\vecs{\theta}^*) \nonumber \\
% &= \log p([\mathbf{y}_i ; \mathbf{y}_j]|[X; X],\vecs{\theta}^*) \\ 
% & \hskip 5pt - \log p(\mathbf{y}_i|X,\vecs{\theta}^*) - \log p(\mathbf{y}_j|X,\vecs{\theta}^*).
%
  & = \frac{1}{2} 
    \begin{bmatrix}
    \mathbf{y}_i^T & \mathbf{y}_j^T
    \end{bmatrix} 
    \left (
    \begin{bmatrix}
      K_y & 0  \\
      0 & K_y
    \end{bmatrix}^{-1} -
    \begin{bmatrix}
      K_y & K  \\
      K & K_y
    \end{bmatrix}^{-1}
    \right )
    \begin{bmatrix}
    \mathbf{y}_i \\ 
    \mathbf{y}_j
    \end{bmatrix}  \nonumber \\
    & \hskip 10pt -\frac{1}{2} \log \frac{\det Q}{\det K_y},
% \end{split}
\label{eq:bayesfactor2}
\end{align}
where $Q$ is the Schur complement:
\begin{align}
\label{eq:Schur}
Q=K_y-K K_y^{-1} K.
\end{align}

Note that the likelihood ratio~\eqref{eq:bayesfactor1} has the same form as the Bayes factor~\citep{kass1995bayes} in Bayesian model selection. From this perspective, our measure can be understood as the comparison between two models in two ways: on one hand, $s( \mathbf{y}_i, \mathbf{y}_j)$ compares modelling two time courses with one single function of the GP versus modelling them with two independent functions of the GP; 
%which can be seen from~\eqref{eq:bayesfactor2}. 
alternatively, it is easy to see that our measure~\eqref{eq:bayesfactor1} can be rewritten as
\begin{align*}
    s( \mathbf{y}_i, \mathbf{y}_j) &= \log p(\mathbf{y}_j | \mathbf{y}_i,X,\vecs{\theta}^*) - \log p(\mathbf{y}_j|X,\vecs{\theta}^*) \\
    &= \log p(\mathbf{y}_i | \mathbf{y}_j,X,\vecs{\theta}^*) - \log p(\mathbf{y}_i|X,\vecs{\theta}^*)
\end{align*}
which is the difference between the log-likelihood of $\mathbf{y}_j$ based on the posterior GP given $\mathbf{y}_i$ compared to the prior GP without $\mathbf{y}_i$ being given. Hence the measure quantifies the improvement in the prediction of $\mathbf{y}_j$ that can be drawn by knowing $\mathbf{y}_i$.  
The measure is symmetric, so the same applies by exchanging $\mathbf{y}_j$ for $\mathbf{y}_i$.

\subsection{Euclidean distance as a limit of the GP log-likelihood ratio}

It can be shown that the Euclidean distance~\eqref{eq:Euclidean} stems naturally from the GP similarity~\eqref{eq:bayesfactor1} in the limit when the noise variance $\sigma_n^2$ is much smaller than the signal variance $\sigma_f^2$.

To see this, recall the Neumann series~\citep{stewart1998matrix} 
%for $(I+\sigma_n^2 K^{-1})^{-1}$ 
%to get 
%\the expansion %for $(K+\sigma_n^2 I)^{-1}$ 
\begin{align*}
   & K_y^{-1}= (K+\sigma_n^2 I)^{-1} = 
   K^{-1} \sum_{m=0}^\infty (-1)^m \left(\sigma_n^2 K^{-1}\right)^m .
%  \\ & = K^{-1} - \sigma_n^2 K^{-2} + \sigma_n^4 K^{-3} - \sigma_n^6 K^{-4} + \cdots
%  \\  &=K^{-1} \Big ( I -\sigma_n^2 K^{-1} + (\sigma_n^2 K^{-1})^2 - (\sigma_n^2 K^{-1})^3 + \cdots \Big ).
\end{align*}
%which exists if $\Vert \sigma_n^2 K^{-1} \Vert < 1$.
%
Noting that the Gaussian kernel matrix $G$ is positive definite, we have that
%if ${\sigma_n^2}/{\sigma_f^2} \to 0$, then 
\begin{equation*}
\text{if } {\sigma_n^2}/{\sigma_f^2} \to 0   \implies
\Vert \sigma_n^2 K^{-1} \Vert =  (\sigma_n^2/\sigma_f^2) \ \Vert G^{-1} \Vert \to 0,
    \label{eq:sigma_k}
\end{equation*}
For small $\sigma_n^2/\sigma_f^2$, we thus take the first two terms of the expansion to~$\mathcal{O}\left( \Vert \sigma_n^2 K^{-1} \Vert^2 \right)$: 
\begin{align}
    \label{eq:approx}
    K K_y^{-1} = K_y^{-1} K  \simeq %K^{-1}  - \sigma_n^2 K^{-2}.
   I -\sigma_n^2 K^{-1}, 
\end{align}
and the Schur complement~\eqref{eq:Schur} is approximated as:
\begin{align*}
& Q \simeq K_y-(I -\sigma_n^2 K^{-1}) K = 2 \sigma_n^2 \, 
\Big (I -\frac{1}{2} \sigma_n^2 K^{-1} \Big), \\
& Q^{-1} \simeq \frac{1} {2 \sigma_n^2} \, \Big (I +\frac{1}{2} \sigma_n^2 K^{-1} \Big).
\end{align*}

To approximate the GP similarity measure~\eqref{eq:bayesfactor2}, we use block matrix inversion:
\begin{align*}
&\begin{bmatrix}
      K_y & 0\\
    0 & K_y
    \end{bmatrix}^{-1} -
\begin{bmatrix}
      K_y & K \\
    K & K_y
    \end{bmatrix}^{-1} \\ 
    & = \begin{bmatrix}
      -K_y^{-1}K Q^{-1} K K_y^{-1} & Q^{-1} K K_y^{-1}\\
    K_y^{-1} K Q^{-1} & -K_y^{-1}K Q^{-1} K K_y^{-1}
    \end{bmatrix} \\
    & \simeq \frac{1}{2 \sigma_n^2} 
    \begin{bmatrix}
      -(I-3\sigma_n^2 K^{-1}/2) & I-\sigma_n^2 K^{-1}/2\\
    I-\sigma_n^2 K^{-1}/2 & -(I-3\sigma_n^2 K^{-1}/2)
    \end{bmatrix} \\
    %
    % & = \frac{-1}{2 \sigma_n^2} \Bigg (
    % \begin{bmatrix}
    %   I & -I \\
    % -I & I
    % \end{bmatrix} - \frac{1}{2}
    %  \begin{bmatrix}
    %   3 \sigma_n^2 K^{-1} & -\sigma_n^2 K^{-1}\\
    % -\sigma_n^2 K^{-1} & 3\sigma_n^2 K^{-1}
    % \end{bmatrix}  \Bigg ) \\
    %
    & = \frac{-1}{2 \sigma_n^2} \Bigg (
    \begin{bmatrix}
      1 & -1 \\
    -1 & 1
    \end{bmatrix} \otimes I - \frac{1}{2}
     \begin{bmatrix}
       3 & -1\\
    -1 & 3
    \end{bmatrix} \otimes \left(\sigma_n^2 K^{-1}\right) \Bigg ) 
    .
\end{align*}
and we approximate the determinant:
\begin{align*}
    \det Q &= \det K_y \, \det \left(I - (K K_y^{-1})^2 \right) \\
    & \simeq \det K_y \, \det \left( 2 \, \sigma_n^2 K^{-1} \right),
\end{align*}
whence we obtain:
\begin{align*}  
& s(\mathbf{y}_i,\mathbf{y}_j)
 \simeq \\  
 &\frac{-1}{4 \sigma_n^2} 
    \begin{bmatrix}
    \mathbf{y}_i^T & \mathbf{y}_j^T
    \end{bmatrix} 
%%%%%
%  \Bigg (
%     \begin{bmatrix}
%       I & -I \\
%     -I & I
%     \end{bmatrix} 
%  - \frac{1}{2}
%      \begin{bmatrix}
%       3 \sigma_n^2 K^{-1} & -\sigma_n^2 K^{-1}\\
%     -\sigma_n^2 K^{-1} & 3\sigma_n^2 K^{-1}
%     \end{bmatrix}  \Bigg )
%%%% Another way of writing the same
\Bigg (
    \begin{bmatrix}
      1 & -1 \\
    -1 & 1
    \end{bmatrix} \otimes I - 
    \frac{\sigma_n^2}{2 \sigma_f^2}
     \begin{bmatrix}
       3 & -1\\
    -1 & 3
    \end{bmatrix} \otimes G^{-1} \Bigg ) 
    \begin{bmatrix}
    \mathbf{y}_i \\ 
    \mathbf{y}_j
    \end{bmatrix}  \nonumber \\
    & \hskip 25pt -\frac{1}{2} \log \det \left(2\, 
    \frac{ \sigma_n^2}{\sigma_f^2} \, G^{-1} \right).
%     \simeq
% \frac{1}{2\sigma_n^2}
%     \begin{bmatrix}
%     I  & -I \\
%     -I  & I 
%     \end{bmatrix} + \frac{1}{2}
%     \begin{bmatrix}
%     0 & K^{-1} \\
%     K^{-1} & 0 
%     \end{bmatrix}.
\end{align*}
Here $\otimes$ denotes the Kronecker product and $G $ is the Gaussian kernel matrix in~\eqref{eq:covariance_matrix}.
The approximation of the GP similarity measure~\eqref{eq:bayesfactor2} to first order is then :
\begin{align}
\label{eq:GP_equals_Euclidean}
s(\mathbf{y}_i,\mathbf{y}_j) \simeq \frac{-1}{4\sigma_n^2} 
\left (\Vert \mathbf{y}_i - \mathbf{y}_j\Vert^2 + 
%%%%%
%\mathcal{O}\left(\Vert\sigma_n^2 K^{-1} \Vert \right) 
%%%%%  Another option...
\mathcal{O}\left(\frac{\sigma_n^2}{\sigma_f^2} \left \Vert G^{-1} \right \Vert \right)
\right ).
% - C \propto -(\mathbf{y}_i^T \mathbf{y}_i - 2\mathbf{y}_i^T\mathbf{y}_j + \mathbf{y}_j^T\mathbf{y}_j) = -\Vert \mathbf{y}_i - \mathbf{y}_j\Vert^2.
\end{align}
Therefore, if ${\sigma_n^2}/{\sigma_f^2} \to 0$ (i.e., when the variance of the noise is much smaller than the signal variance), the  dissimilarity measure $-s(\mathbf{y}_i,\mathbf{y}_j)$ is equivalent to the Euclidean distance $\Vert \mathbf{y}_i - \mathbf{y}_j\Vert^2$.
Note that this relationship holds not only for the Gaussian kernel $G(x,x')$ but for any positive definite kernel. 

\subsection{Asynchronous time courses}

Although, for simplicity of exposition, we have concentrated on the case of synchronous time sampling, the GP similarity measure is equally applicable to non-synchronous samples. In our derivations above, synchronous time points are only necessary to obtain the formal limit to the Euclidean distance in Eq.~\ref{eq:GP_equals_Euclidean}. 

To see the applicability to non-synchronous samples, note that the likelihoods in Eq.~\ref{eq:likelihood_same} and Eq.~\ref{eq:likelihood_diff} do not require the time courses to have the same time points. We can therefore consider $N$ time courses denoted by $\mathbf{y}_i = [y_{i,1},\cdots,y_{i,t_i}]$ of length $t_i$, sampled at (potentially) distinct time points $X_i = [x_{i,1},\cdots,x_{i,t_i}]$ ($i=1,\ldots,N$). In this case, we can similarly model the time courses with GP and learn the hyperparameters by maximising
\begin{equation}
\label{eq:inference_async}
    \vecs{\theta}^* = \argmax_{\vecs{\theta}} \sum_{i=1}^N \log p(\mathbf{y_i}|X_i,\vecs{\theta}),
\end{equation}
and it is then easy to write the two log-likelihoods: 
%$\log p_\text{diff}(\mathbf{y}_i,\mathbf{y}_j|\vecs{\theta}^*)$ and $\log p_\text{same}(\mathbf{y}_i,\mathbf{y}_j|\vecs{\theta}^*) $ as,
\begin{align}
 %   \begin{split}
        & \log p_\text{diff}(\mathbf{y}_i,\mathbf{y}_j|\vecs{\theta}^*) =
        \log p(\mathbf{y}_i|X_i,\vecs{\theta}^*) + \log p(\mathbf{y}_j|X_j,\vecs{\theta}^*) \nonumber \\
       & \hskip 10pt = -\frac{1}{2}
       \begin{bmatrix}
    \mathbf{y}_i^T & \mathbf{y}_j^T
    \end{bmatrix} 
    \begin{bmatrix}
      K_{y_i} & 0  \nonumber \\
    0 & K_{y_j}
    \end{bmatrix}^{-1}
    \begin{bmatrix}
    \mathbf{y}_i \\ 
    \mathbf{y}_j
    \end{bmatrix} \\
    & \hskip 20pt -\frac{1}{2} \log \det
    \begin{bmatrix}
    K_{y_i} & 0 \\
    0 & K_{y_j} 
    \end{bmatrix}
    - \frac{1}{2}(t_i+t_j) \, \log 2\pi,
 %   \end{split}
 \label{eq:likelihood_diff_async}
\end{align}
\begin{equation}
    \label{eq:likelihood_same_async}
\begin{split}
    &\log p_\text{same}(\mathbf{y}_i,\mathbf{y}_j|\vecs{\theta}^*) = \log p\left(
    \begin{bmatrix}
    \mathbf{y}_i \\ 
    \mathbf{y}_j
    \end{bmatrix}\bigg\vert 
    \begin{bmatrix}
    X_i \\ 
    X_j
    \end{bmatrix},\vecs{\theta}^*\right) \\ 
    & \hskip 10pt = - \frac{1}{2} 
    \begin{bmatrix}
    \mathbf{y}_i^T & \mathbf{y}_j^T
    \end{bmatrix} 
    \begin{bmatrix}
      K_{y_i} & K_{ij}  \\
    K_{ij}^T & K_{y_j}
    \end{bmatrix}^{-1}
    \begin{bmatrix}
    \mathbf{y}_i \\ 
    \mathbf{y}_j
    \end{bmatrix} \\
    & \hskip 20pt -\frac{1}{2} \log \det
    \begin{bmatrix}
    K_{y_i} & K_{ij}\\
    K_{ij}^T & K_{y_j} 
    \end{bmatrix}
    -  \frac{1}{2}(t_i+t_j) \, \log 2\pi.
\end{split}
\end{equation}
where $K_{y_i}$ and $K_{y_j}$ are the covariance matrices of $\mathbf{y}_i$ and $\mathbf{y}_j$ with sizes $t_i \times t_i$ and $t_j \times t_j$, respectively; and $K_{ij}$ is the cross-covariance matrix between $\mathbf{y}_i$ and $\mathbf{y}_j$ with size $t_i \times t_j$. The GP similarity between the two time courses can again be computed as the difference between the two log-likelihoods~\eqref{eq:likelihood_diff_async}~and~\eqref{eq:likelihood_same_async}. 

\subsection{Computational complexity}
In terms of computational complexity, fitting a GP model with $N$ time courses of length $t$ takes $\mathcal{O}(t^3+Nt^2)$ time. Computing pairwise similarities takes $\mathcal{O}(tN^2)$ time. Since we deal with high-dimensional short time courses ($N \gg t$), the total time for GP similarity would be approximately $\mathcal{O}(tN^2)$, which is the same as for the Euclidean distance. 

Extra computational time is needed for the asynchronised time courses, where all the time courses have a different covariance matrix. It results in a computational time of $\mathcal{O}(Nt^3)$ for model fitting and $\mathcal{O}(N^2t^3)$ for computing the pairwise GP similarity if all the time courses are of average length $t$. This might be a limitation to applications with large $N$ and $t$.

\section{Numerical experiments}
%We have shown that if ${\sigma_n^2}/{\sigma_f^2} $ approaches zero, using the proposed similarity measure is the same as the Euclidean distance. However, in practice, it is very rare that the observed data is noiseless. After optimising the log-marginal likelihood the inferred $\sigma_n$ is usually not negligible compared to $\sigma_f$. To 
To test the applicability of the proposed GP similarity measure, we have run numerical experiments on synthetic and real sparse time course data. Our results show that the similarity $s(\mathbf{y}_i,\mathbf{y}_j)$ is more robust to observational noise than the Euclidean distance when used as the basis to cluster time courses using two standard clustering algorithms (hierarchical and spectral clustering). We also compare the performance of the GP similarity against the Bregman divergence and Dynamic Time Warping.
%in which the similarity measure is an important ingredient. We validate the performance by clustering of both synthetic and real data.

\subsection{Synthetic data}
\begin{figure}[!b]
    \centering
    \includegraphics[width=0.9\linewidth]{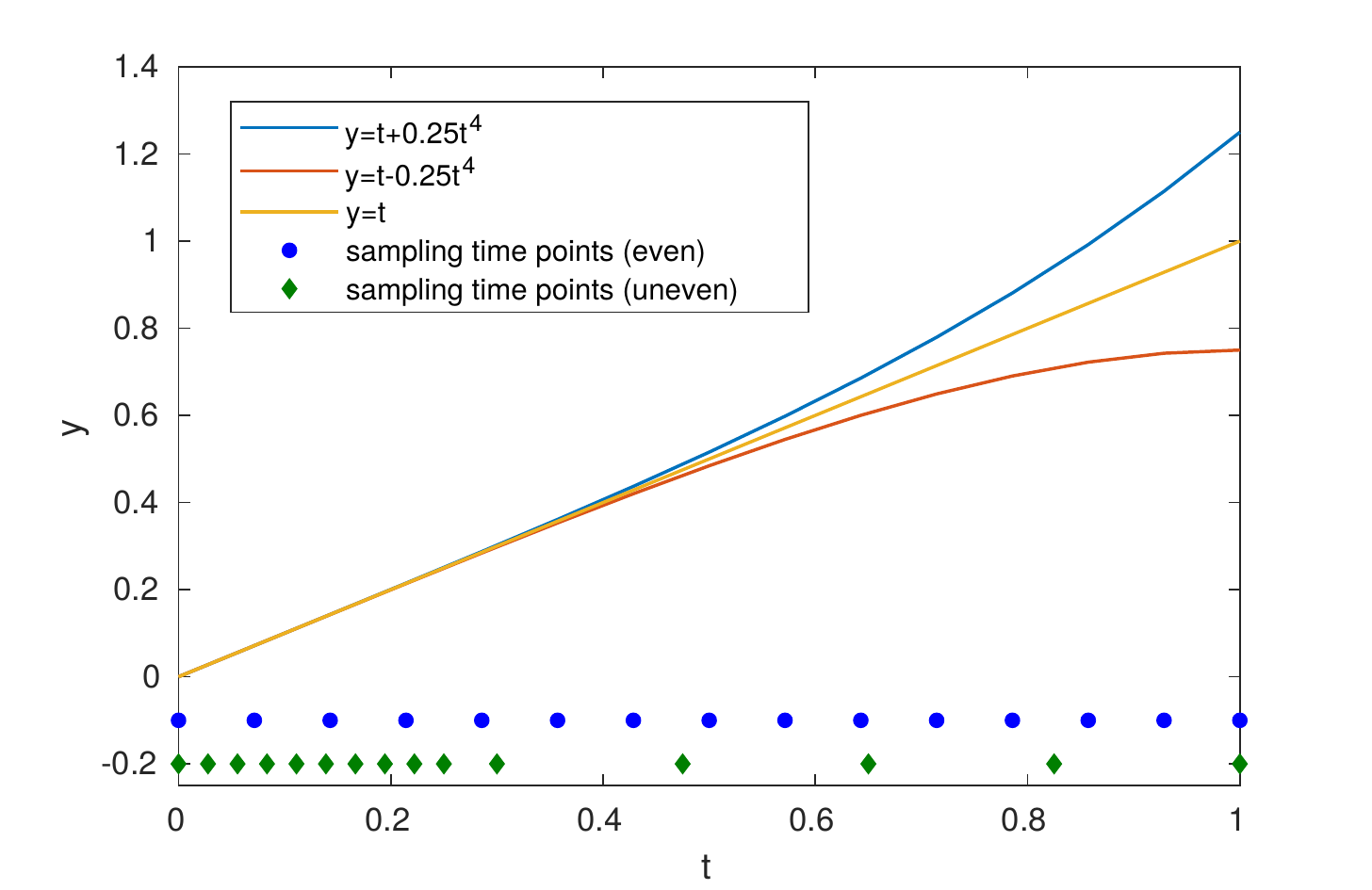}
    \caption{The three functions and the sampling time positions used to generate the synthetic data. The results of even sampling are in Fig.~\ref{fig:even}. The results of uneven sampling are in Fig.~\ref{fig:uneven}.}
    \label{fig:3functions}
\end{figure}

Our synthetic dataset is obtained by sampling from the three different time profiles shown in Fig.~\ref{fig:3functions} with additive Gaussian noise. 
From each of the three time profiles, we generate 50 evenly-sampled time courses of length $t=15$ with a given level of sampling noise. Our task is to cluster the 150 time series into 3 groups in an unsupervised manner.
To do this, we compute the pairwise similarities (or distances) between the $150$ time courses using the GP similarity~
\eqref{eq:bayesfactor1}.
%\MB{There are 150 time courses for each level of noise, right??}
From the computed similarity matrix, we then cluster the samples using two well-known methods: (i) spectral clustering with $k$NN graph ($k=7$)~\citep{yu2003multiclass}; (ii) agglomerative hierarchical clustering with average linkage~\citep{rokach2005clustering}.
We then repeat the numerical experiment 100 times in each case. To evaluate the clustering performance, we use the normalised mutual information (NMI)~\citep{vinh2010information} against the known ground truth (i.e., the three profiles used to generate the data). 

We then repeat the same procedure with three popular measures: the Euclidean distance, the Dynamic Time Warping (DTW) distance and the Bregman divergence in the RKHS.
Given that the only varying ingredient is the similarity measure, the clustering performance reflects the quality of the similarity measure for this purpose.

\begin{figure}[!h]
\centering
\subfloat[Noise level = 0.08]{\includegraphics[width=0.95\linewidth]{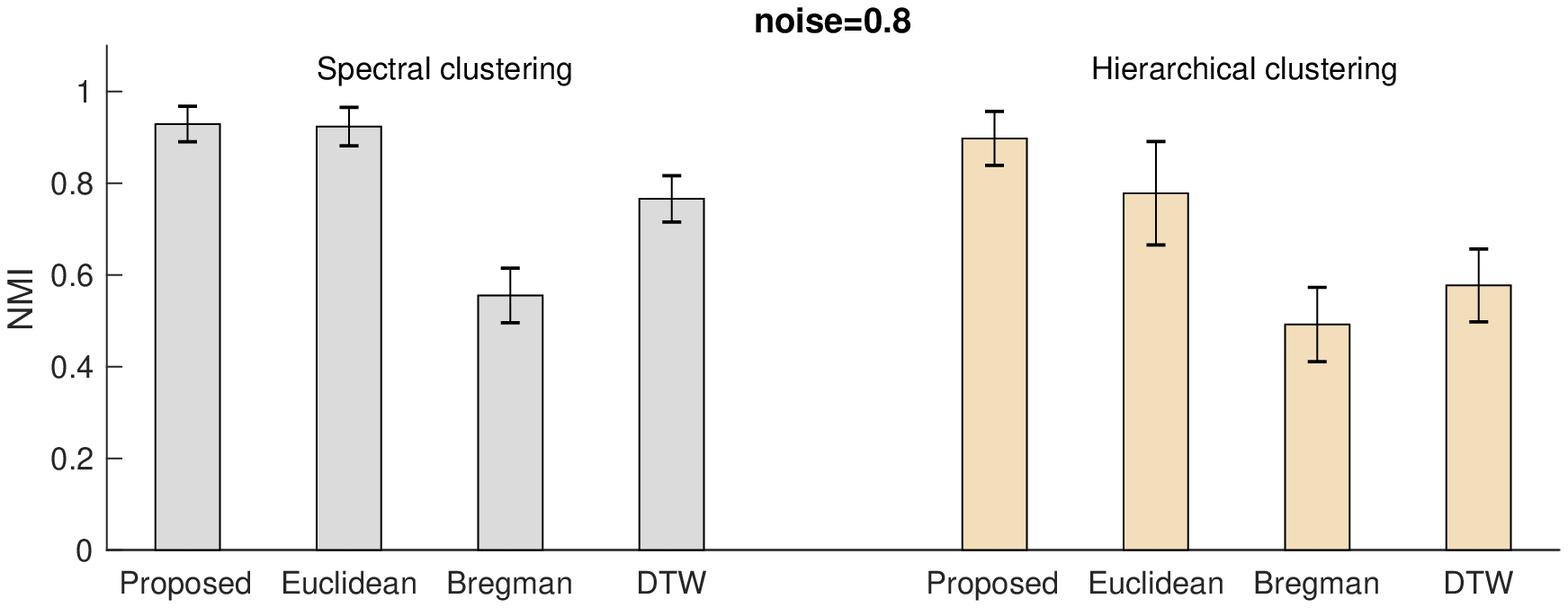}%
\label{fig:even1}}
\hfil
\subfloat[Noise level = 0.10]{\includegraphics[width=0.95\linewidth]{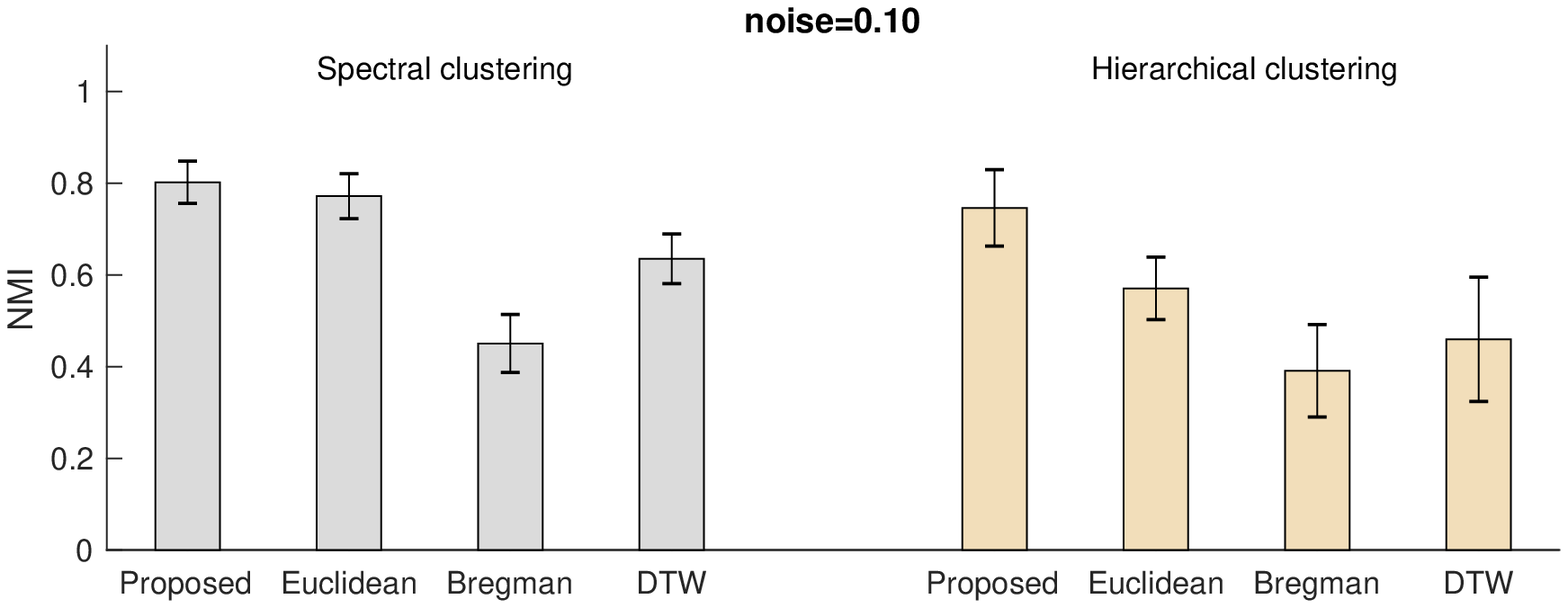}%
\label{fig:even2}}
\hfil
\subfloat[Noise level = 0.12]{\includegraphics[width=0.95\linewidth]{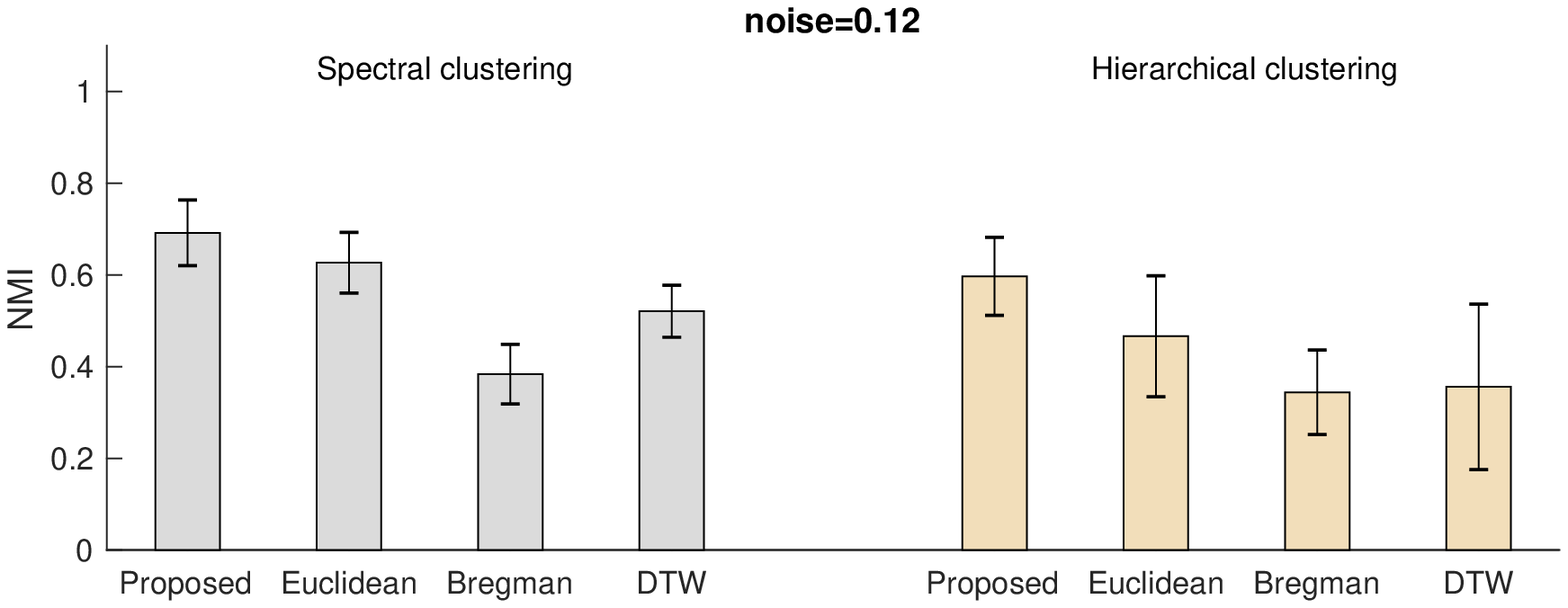}%
\label{fig:even3}}
\caption{Clustering performance (measured as NMI) for spectral clustering with $k$NN and hierarchical clustering based on the proposed GP similarity~\eqref{eq:bayesfactor1}, Euclidean distance~\eqref{eq:Euclidean}, DTW and Bregman divergence in the RKHS with increasing noise levels from (a)-(c). 
The sampling time points are equally spaced between 0 and 1
(see Fig.~\ref{fig:3functions}).}
\label{fig:even}
\end{figure}

Figure~\ref{fig:even} shows the clustering quality (0 $\leq$ NMI $\leq$ 1) for evenly sampled time series with increasing levels of sampling noise achieved with both clustering methods and all four distance/similarity functions. 
As expected from~\eqref{eq:GP_equals_Euclidean}, both Euclidean and GP similarity give comparable results for small sampling noise when using spectral clustering. Note, however, that even for small noise, the performance of the GP similarity is superior to the Euclidean distance when using hierarchical clustering (Fig.~\ref{fig:even1}), a method that is very sensitive to noisy data. Also as expected the DTW performs worse than the Euclidean distance for the synchronous case, since there is no need for alignment. The Bregman divergence does not perform well in this case because it measures the distance between the two continuous curves in the RKHS fitted from the time courses data and does not consider the uncertainty due to noise.

As the observation noise increases, our GP similarity measure gains further advantage over the Euclidean distance for both clustering methods (Figs.~\ref{fig:even2}--\ref{fig:even3}). 
%\MBB{Add somewhere around here your p value statements within the text....}
The p-values for a Wilcoxon rank-sum test between the NMI values of Euclidean and GP similarity with spectral clustering for the noise levels 0.08, 0.10 and 0.12 are 0.043, 2.5e-7 and 7.6e-16, respectively. The p-values for a Wilcoxon rank-sum test between the NMI values of Euclidean and GP similarity with hierarchical clustering for the noise levels 0.08, 0.10 and 0.12 are 1.4e-15, 1.7e-25, and 1.1e-20, respectively. Hence the GP similarity measure performs significantly better than the Euclidean distance for clustering with both algorithms. 

In general, spectral clustering always performs better than hierarchical clustering, which is more sensitive to noise, 
and the best performance is obtained consistently using spectral clustering with GP similarity.
Note that the performance obtained with spectral clustering using other metrics can be achieved at a lower computational cost using hierarchical clustering with GP similarity.

%\MB{For Arxiv, you should change the figures to have 'GP similarity' instead of 'Proposed'.  Has 'GP similarity' been used already?  }

\begin{figure}[!th]
\centering
\subfloat[Noise level = 0.08]{\includegraphics[width=0.99\linewidth]{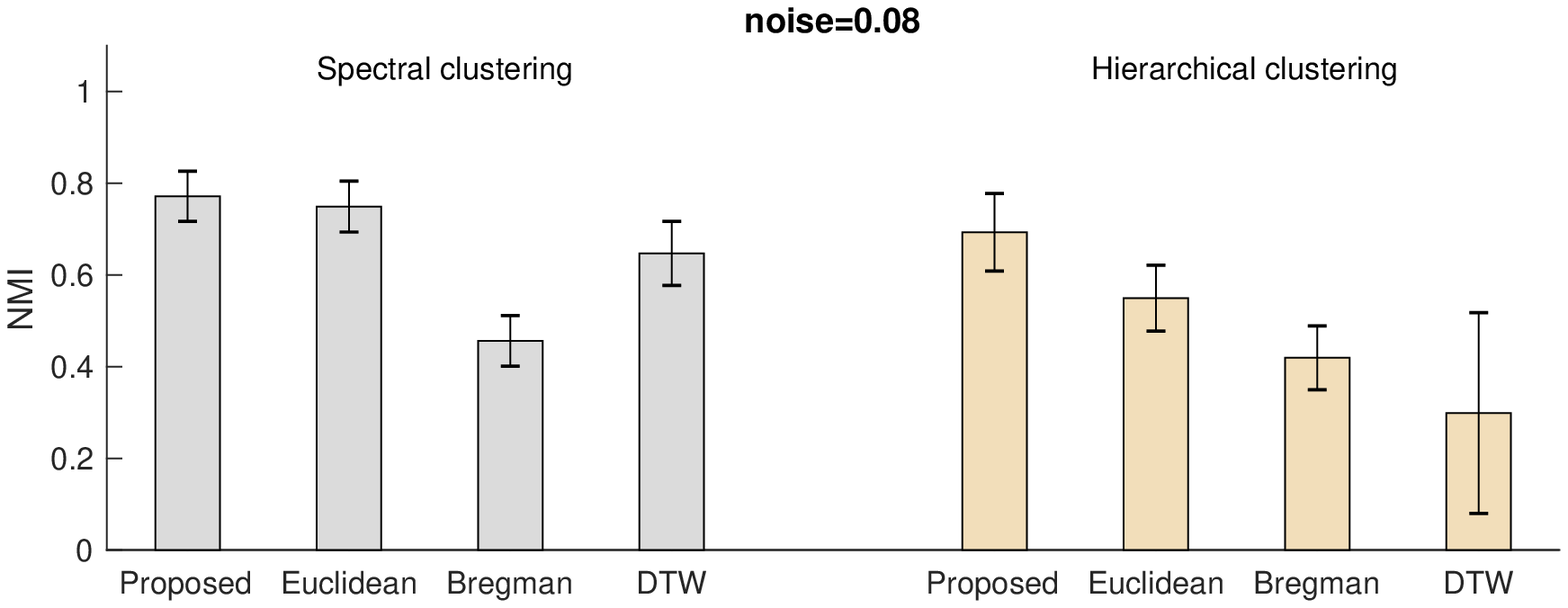}%
\label{fig:uneven1}}
\hfil
\subfloat[Noise level = 0.10]{\includegraphics[width=0.99\linewidth]{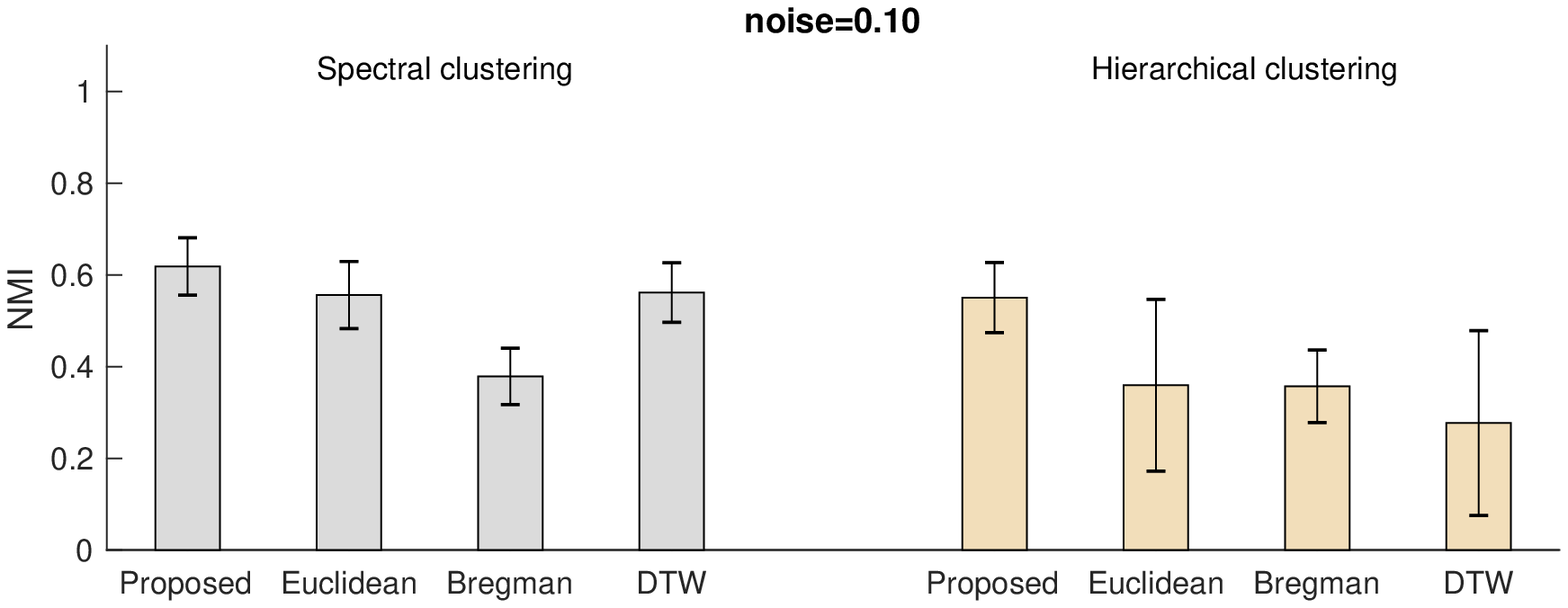}%
\label{fig:uneven2}}
\caption{Clustering performance (measured as NMI) for spectral clustering with $k$NN and hierarchical clustering based on the proposed GP similarity~\eqref{eq:bayesfactor1}, Euclidean distance~\eqref{eq:Euclidean}, DTW and Bregman divergence in the RKHS with increasing noise levels from (a)-(b). In this case, the time points are irregularly sampled between 0 and 1 (see Fig.~\ref{fig:3functions}). The GP similarity consistently outperforms the other distances, especially for larger amounts of noise.}
\label{fig:uneven}
\end{figure}

We also analyse a second set of 150 time series of length $t=15$ collected from the same three functions in Fig.~\ref{fig:3functions} but sampled inhomogeneously in time. The computational procedure is identical to the case of evenly sampled series described above. 
Figure~\ref{fig:uneven} shows that the advantage of the GP similarity measure against the other distances is more prominent when the time points are unevenly sampled. This result highlights the fact that point-to-point similarities (such as the Euclidean distance) can miss important information contained in the long-term trends of the time profiles if the sampling is concentrated irregularly in particular time periods.

\begin{figure}[!th]
\centering
\subfloat[asynchronous time course]{\includegraphics[width=0.5\linewidth]{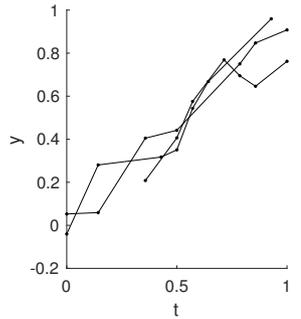}%
\label{fig:async1}}
\hfil
\subfloat[NMI results]{\includegraphics[width=0.9\linewidth]{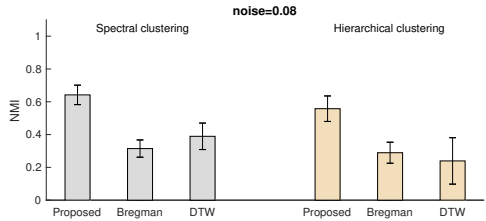}%
\label{fig:async2}}
\caption{(a) Three sample time courses with asyncronised measurements with Noise level = 0.08. (b) Clustering performance (measured as NMI) for spectral clustering with $k$NN and hierarchical clustering based on the proposed GP similarity, DTW and Bregman divergence in the RKHS. The GP similarity consistently outperforms the other distances for both clustering methods.}
\label{fig:async}
\end{figure}

We next tested the non-synchronous case. We again generate a set of 150 time series of length $t=15$ from the same three functions in Fig.~\ref{fig:3functions}. We then make the asynchronous time courses by randomly removing 6,7 or 8 time points from our 15-point time course data (Fig.~\ref{fig:async1}). In this case, the Euclidean distance is not defined. The proposed GP similarity achieves substantially better results than the Bregman divergence and DTW on these examples.

In summary, our numerical experiments on synthetic data indicate improved performance of the GP similarity measure~\eqref{eq:bayesfactor1}. As the observational noise decreases, the performance of the GP similarity measure is equivalent to the Euclidean distance~\eqref{eq:GP_equals_Euclidean}. Although the GP similarity shows enhanced performance for both clustering methods, the improvement is larger for hierarchical clustering, in the presence of large amounts of noise, and under uneven time sampling. As a side comment, we also note that the inclusion of the noise variance $\sigma_n^2$ in the covariance function $K_y$ makes the GP similarity measure more robust for numerical computations, as it reduces numerical instability. 
%For computation, it also guarantees that the covariance matrix $K_y$ is not near singular and improves numerical stability.

\subsection{Application to gene expression time course data}

We next tested the GP similarity measure on a real time course dataset which characterises the gene expressions during the process of cellular reprogramming where differentiated cells are reverted to stem cells~\citep{di2014c}. 
In this dataset, gene expression values were measured at five time points: 0, 2, 4, 6 and 8 days after the cellular reprogramming started. Two biological replicates were measured at each time point. The full measurements include around 40,000 genes. Following standard practice, we select 1912 genes that are highly variable over time. We then follow the same procedure as for the synthetic data above to cluster the 1912 gene expression time courses to extract groups of genes that have similar time profiles during cellular reprogramming.

\begin{figure}[!th]
\centering
\subfloat[Time course of gene \emph{Samd14}]{\includegraphics[width=0.7\linewidth]{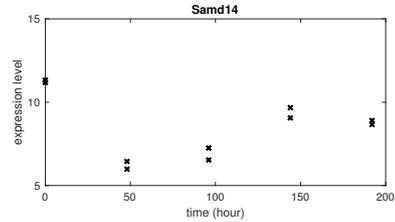}%
\label{fig:g1}}
\hfil
\subfloat[Time course of gene \emph{Fbin5}]{\includegraphics[width=0.7\linewidth]{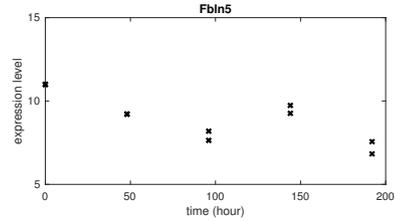}%
\label{fig:g2}}
\hfil
\subfloat[Time course of gene \emph{Serpine1}]{\includegraphics[width=0.7\linewidth]{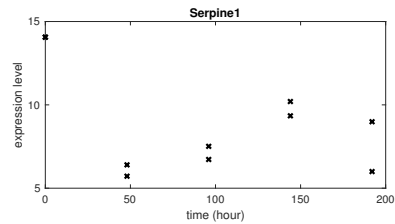}%
\label{fig:g3}}
\caption{The time courses of three genes (each time course was measured on two bioogical replicates). The Euclidean distances from \emph{Samd14} to \emph{Fbin5} and \emph{Serpine1} are both $5.09$. In contrast, the GP dissimilarity ($-s(\mathbf{y}_i,\mathbf{y}_j)$) between \emph{Samd14} and \emph{Fbin5} ($37.48$) is larger than between \emph{Samd14} and \emph{Serpine1} ($25.57$), capturing the fact that the time profile of \emph{Serpine1} is more similar to \emph{Samd14} than to \emph{Fbin5}.}
\label{fig:genes}
\end{figure}

\begin{table*}[!ht]
  \caption{BHI z-scores of the two clustering methods with different number of clusters obtained by analysing the time courses of 1912 highly variable genes in the stem cell transcriptomic dataset of Refs.~\citep{di2014time,di2014c}.}
    \label{tab:BHIs}
    \centering
    \begin{tabular}{|c|c|c|c|c|c|c|}
    \cline{3-7} 
      \multicolumn{2}{c|}{} &   \multicolumn{5}{|c|}{\textit{No.\ of clusters}}  \\ \cline{3-7} \hline
       \textit{Method} 
%        \multirow{2}{*}{\begin{tabular}[c]{@{}c@{}} \textit{Clustering       \\method} \end{tabular}} 
       & \textit{Similarity} & 9 & 11 & 13 & 15 & 16   \\ \hline 
       \multirow{4}{*}{\begin{tabular}[c]{@{}c@{}}Spectral 
       %\\ clustering
       \end{tabular}} & Proposed  & {2.505} & \textbf{4.250} & 1.360 & {3.368} & {3.674} \\ \cline{2-7}
       & Euclidean  & 1.368 & 2.067 & 2.460 & 2.351 & 2.919 \\ \cline{2-7}
       & DTW  & 2.636 & 2.087 & 2.774 & 2.720 & 3.182 \\ \cline{2-7} 
        & Bregman  & {4.229} & {2.725} & 2.652 & {3.617} & {3.520} \\ \hline \hline
       \multirow{4}{*}{\begin{tabular}[c]{@{}c@{}}Hierarchical  
       %\\ clustering
       \end{tabular}} & Proposed  & 1.182 & 2.384 & 2.366 & 1.849 & \textbf{3.353} \\ \cline{2-7} 
       & Euclidean  & -0.090 & 0.074 & -0.266 & -0.215 & -0.319 \\ \cline{2-7} 
       & DTW  & 0.542 & 0.566 & 0.499 & 0.735 & 0.752 \\ \cline{2-7} 
       & Bregman  & 1.392 & 1.303 & 1.139 & 1.388 & 1.508 \\
       \hline
    \end{tabular}
\end{table*}

For this dataset, there is no known `ground truth' against which to compare the obtained clusters. To assess the quality of the clustering, we use a biological score: a modified version of the biological homogeneity index (BHI). The BHI measures within-cluster homogeneity of the genes in terms of biological knowledge and is given by:
\begin{align*}
    \text{BHI} = \frac{1}{L} \sum_{l=1}^L \frac{1}{n_l(n_l-1)} \sum_{\substack{
    i \neq j\\
    i,j\in \mathrm{cluster}\, l }}
    \mathcal{S}(i,j),
\end{align*}
where $L$ is the number of clusters, $n_l$ is the size of cluster $l$, and $\mathcal{S}(i,j)$ is a \textit{biological similarity} between gene $i$ and gene $j$.
In the original BHI~\citep{datta2006methods}, the similarity between genes is either $0$ or $1$, which is an indicator function of two genes sharing any Gene Ontology (GO) terms. In order to better capture the biological information, we use an information-theoretic semantic gene similarity based on Gene Ontologies~\citep{resnik1999semantic,lord2003investigating,yu2010gosemsim}. 
%Since the semantic similarity is built from the perspective of information theory, the
%modified 
%The BHI score of the dataset is compared against the 
For a given clustering, we compute the BHI of $1000$ random clusterings with the same number and size of clusters.
%If the identified clustering has a significantly higher BHI than the random clustering, it is evidence that the clustering is able to capture biological meaningful information. We compute 
We then use the z-score of the BHI against the random clusterings as a level of significance for the obtained clustering.

Since the number of clusters is unknown, we compute the z-score of the BHI for clusterings with different numbers of clusters (Table~\ref{tab:BHIs}).
Again, we find that spectral clustering with GP similarity achieves the best performance. On the other extreme, hierarchical clustering based on Euclidean distances performs no better than random clustering, an indication that the gene expression data has high levels of noise $\sigma_n$. As was the case for the synthetic data above, using GP similarity improves the performance of hierarchical clustering substantially, almost to a comparable level to the spectral method (but below). These results underscore the ability of the GP similarity to deal with noisy data. 

To illustrate visually the behaviour of the GP similarity, Figure~\ref{fig:genes} shows the time courses of three genes where the GP similarity and Euclidean distance behave differently.
In our chosen examples, the time course of the \emph{Samd14} gene (Fig.~\ref{fig:g1}) has the same Euclidean distance to both the \emph{Fbin5} gene (Fig.~\ref{fig:g2}) and the \emph{Serpine1} gene (Fig.~\ref{fig:g3}). On the other hand, according to our GP similarity, the \emph{Serpine1} time course is more similar to \emph{Samd14} than to \emph{Fbin5}, in accordance with our visual expectation from the observed time course profiles. The codes and data are available online in \url{https://github.com/barahona-research-group/BayesFactorSimilarity}.

\section{Conclusion}

In this paper, we have proposed a similarity measure for sparse time course data based on Gaussian processes.
%GPs have been previously applied to gene expression time series to detect differentially expressed genes~\citep{stegle2010robust,kalaitzis2011simple} and to infer the dynamics of transcriptional regulation~\citep{lawrence2007modelling,gao2008gaussian}.
Modelling the time courses with a GP, we use the difference between two log-likelihoods (in the form of a Bayes factor) as a GP similarity measure. We show that the proposed measure is equivalent to the Euclidean distance in the limit where the noise variance in the observations is negligible compared to the signal variance. The proposed measure is computationally simple and can be easily extended to the cases when the time courses are not synchronously observed at the same time points.

Our numerical experiments show that the proposed measure has improved robustness to noise when used for data clustering with different clustering methods. The advantage of the proposed measure over the Euclidean distance is more noticeable with hierarchical clustering, under high noise, and with uneven time sampling. 

We note two limitations of the proposed measure. First, the GP similarity is not a metric. Indeed, $s(\mathbf{y}_i,\mathbf{y}_i)$ is not zero but instead, it gives an estimate of the level of noise in $\mathbf{y}_i$. %Of course, if we do not consider any noise in the model, the proposed measure becomes a metric, since it is then equivalent to the Euclidean distance. 
Second, the high computational cost for the non-synchronous data may be a limitation if both the length $t$ and number of time courses $N$ are large. Sparse GP models can be used to overcome this limitation~\citep{liu2020gaussian}.
Although in this study we have only considered a GP over time as a one-dimensional space, future studies could generalise the GP similarity measure to data that can be modelled by GPs in a latent space, or on any other smooth manifold, further enhancing its applicability.

%\citep{schulam2016disease} \citep{futoma2016predicting}
%Comparison to papers by(Schulam 2016, Futoma 2016): Those papers have a distinct focus on dimensionality reduction of EHR data using latent models. Using our similarity for dimensionality reduction would be an interesting direction in future. For instance, one can infer the model with the methods in those papers, and then use our measure to compute the similarity between time series for dimensionality reduction.

\begin{acknowledgements} % will be removed in pdf for initial submission,
                         % so you can already fill it to test with the
                         % ‘accepted’ class option
    This work was supported by the European Commission [European Union 7th Framework Programme for research, technological development and demonstration under grant agreement no. 607466]; and the Engineering and Physical Sciences Research Council [under grant EP/N014529/1 to M.B. funding the EPSRC Centre for Mathematics of Precision Healthcare].
\end{acknowledgements}

\bibliography{refs}

\begin{thebibliography}{36}
\providecommand{\natexlab}[1]{#1}
\providecommand{\url}[1]{\texttt{#1}}
\expandafter\ifx\csname urlstyle\endcsname\relax
  \providecommand{\doi}[1]{doi: #1}\else
  \providecommand{\doi}{doi: \begingroup \urlstyle{rm}\Url}\fi

\bibitem[Barahona and Poon(1996)]{barahona1996}
Mauricio Barahona and Chi-Sang Poon.
\newblock Detection of nonlinear dynamics in short, noisy time series.
\newblock \emph{Nature}, 381\penalty0 (6579):\penalty0 215--217, 1996.

\bibitem[Brillinger(1981)]{brillinger1981time}
David~R Brillinger.
\newblock \emph{Time series: data analysis and theory}, volume~36.
\newblock Siam, 1981.

\bibitem[Chatfield(2003)]{chatfield2003analysis}
Chris Chatfield.
\newblock \emph{The analysis of time series: an introduction}.
\newblock Chapman and Hall/CRC, 2003.

\bibitem[Chen et~al.(2005)Chen, \"{O}zsu, and Oria]{chen2005robust}
Lei Chen, M.~Tamer \"{O}zsu, and Vincent Oria.
\newblock Robust and fast similarity search for moving object trajectories.
\newblock In \emph{Proceedings of the 2005 ACM SIGMOD International Conference
  on Management of Data}, SIGMOD '05, pages 491--502, New York, NY, USA, 2005.
  ACM.
\newblock ISBN 1-59593-060-4.
\newblock \doi{10.1145/1066157.1066213}.

\bibitem[Datta and Datta(2006)]{datta2006methods}
Susmita Datta and Somnath Datta.
\newblock Methods for evaluating clustering algorithms for gene expression data
  using a reference set of functional classes.
\newblock \emph{BMC bioinformatics}, 7\penalty0 (1):\penalty0 397, 2006.

\bibitem[Di~Stefano et~al.(2014{\natexlab{a}})Di~Stefano, Collombet, and
  Graf]{di2014time}
Bruno Di~Stefano, Samuel Collombet, and Thomas Graf.
\newblock Time-resolved gene expression profiling during reprogramming of
  {C/EBP}$\alpha$-pulsed {B} cells into {iPS} cells.
\newblock \emph{Scientific data}, 1, 2014{\natexlab{a}}.

\bibitem[Di~Stefano et~al.(2014{\natexlab{b}})Di~Stefano, Sardina, van Oevelen,
  Collombet, Kallin, Vicent, Lu, Thieffry, Beato, and Graf]{di2014c}
Bruno Di~Stefano, Jose~Luis Sardina, Chris van Oevelen, Samuel Collombet,
  Eric~M Kallin, Guillermo~P Vicent, Jun Lu, Denis Thieffry, Miguel Beato, and
  Thomas Graf.
\newblock C/ebp$\alpha$ poises {B} cells for rapid reprogramming into induced
  pluripotent stem cells.
\newblock \emph{Nature}, 506\penalty0 (7487):\penalty0 235, 2014{\natexlab{b}}.

\bibitem[Fan et~al.(2011)Fan, Lv, and Qi]{fan2011sparse}
Jianqing Fan, Jinchi Lv, and Lei Qi.
\newblock Sparse high-dimensional models in economics.
\newblock \emph{Annu. Rev. Econ.}, 3\penalty0 (1):\penalty0 291--317, 2011.

\bibitem[Fulcher and Jones(2014)]{fulcher2014highly}
Ben~D Fulcher and Nick~S Jones.
\newblock Highly comparative feature-based time-series classification.
\newblock \emph{IEEE Transactions on Knowledge and Data Engineering},
  26\penalty0 (12):\penalty0 3026--3037, 2014.

\bibitem[Gao et~al.(2008)Gao, Honkela, Rattray, and Lawrence]{gao2008gaussian}
Pei Gao, Antti Honkela, Magnus Rattray, and Neil~D Lawrence.
\newblock Gaussian process modelling of latent chemical species: applications
  to inferring transcription factor activities.
\newblock \emph{Bioinformatics}, 24\penalty0 (16):\penalty0 i70--i75, 2008.

\bibitem[G{\'o}recki(2014)]{gorecki2014using}
Tomasz G{\'o}recki.
\newblock Using derivatives in a longest common subsequence dissimilarity
  measure for time series classification.
\newblock \emph{Pattern Recognition Letters}, 45:\penalty0 99--105, 2014.

\bibitem[Hedeker and Gibbons(2006)]{hedeker2006longitudinal}
Donald Hedeker and Robert~D Gibbons.
\newblock \emph{Longitudinal data analysis}, volume 451.
\newblock John Wiley \& Sons, 2006.

\bibitem[Kalaitzis and Lawrence(2011)]{kalaitzis2011simple}
Alfredo~A Kalaitzis and Neil~D Lawrence.
\newblock A simple approach to ranking differentially expressed gene expression
  time courses through {Gaussian} process regression.
\newblock \emph{BMC bioinformatics}, 12\penalty0 (1):\penalty0 1, 2011.

\bibitem[Kass and Raftery(1995)]{kass1995bayes}
Robert~E Kass and Adrian~E Raftery.
\newblock Bayes factors.
\newblock \emph{Journal of the American Statistical Association}, 90\penalty0
  (430):\penalty0 773--795, 1995.

\bibitem[Kayano et~al.(2016)Kayano, Matsui, Yamaguchi, Imoto, and
  Miyano]{kayano2016gene}
Mitsunori Kayano, Hidetoshi Matsui, Rui Yamaguchi, Seiya Imoto, and Satoru
  Miyano.
\newblock Gene set differential analysis of time course expression profiles via
  sparse estimation in functional logistic model with application to
  time-dependent biomarker detection.
\newblock \emph{Biostatistics}, 17\penalty0 (2):\penalty0 235--248, 2016.

\bibitem[Keogh and Kasetty(2003)]{keogh2003need}
Eamonn Keogh and Shruti Kasetty.
\newblock On the need for time series data mining benchmarks: a survey and
  empirical demonstration.
\newblock \emph{Data Mining and knowledge discovery}, 7\penalty0 (4):\penalty0
  349--371, 2003.

\bibitem[Keogh and Ratanamahatana(2005)]{keogh2005exact}
Eamonn Keogh and Chotirat~Ann Ratanamahatana.
\newblock Exact indexing of dynamic time warping.
\newblock \emph{Knowledge and Information Systems}, 7\penalty0 (3):\penalty0
  358--386, 2005.

\bibitem[Lawrence et~al.(2007)Lawrence, Sanguinetti, and
  Rattray]{lawrence2007modelling}
Neil~D. Lawrence, Guido Sanguinetti, and Magnus Rattray.
\newblock Modelling transcriptional regulation using gaussian processes.
\newblock In B.~Sch\"{o}lkopf, J.~C. Platt, and T.~Hoffman, editors,
  \emph{Advances in Neural Information Processing Systems 19}, pages 785--792.
  MIT Press, 2007.

\bibitem[Liao(2005)]{liao2005clustering}
T~Warren Liao.
\newblock Clustering of time series data—a survey.
\newblock \emph{Pattern recognition}, 38\penalty0 (11):\penalty0 1857--1874,
  2005.

\bibitem[Liu et~al.(2020)Liu, Ong, Shen, and Cai]{liu2020gaussian}
Haitao Liu, Yew-Soon Ong, Xiaobo Shen, and Jianfei Cai.
\newblock When gaussian process meets big data: A review of scalable gps.
\newblock \emph{IEEE transactions on neural networks and learning systems},
  31\penalty0 (11):\penalty0 4405--4423, 2020.

\bibitem[Lord et~al.(2003)Lord, Stevens, Brass, and
  Goble]{lord2003investigating}
Phillip~W. Lord, Robert~D. Stevens, Andy Brass, and Carole~A. Goble.
\newblock Investigating semantic similarity measures across the gene ontology:
  the relationship between sequence and annotation.
\newblock \emph{Bioinformatics}, 19\penalty0 (10):\penalty0 1275--1283, 2003.

\bibitem[Lu et~al.(2008)Lu, Leen, Huang, and Erdogmus]{lu2008reproducing}
Zhengdong Lu, Todd~K. Leen, Yonghong Huang, and Deniz Erdogmus.
\newblock A reproducing kernel hilbert space framework for pairwise time series
  distances.
\newblock In \emph{Proceedings of the 25th International Conference on Machine
  Learning}, ICML ’08, page 624–631, New York, NY, USA, 2008. Association
  for Computing Machinery.
\newblock ISBN 9781605582054.
\newblock \doi{10.1145/1390156.1390235}.
\newblock URL \url{https://doi.org/10.1145/1390156.1390235}.

\bibitem[Ndukum et~al.(2011)Ndukum, Fonseca, Santos, Voit, and
  Datta]{ndukum2011statistical}
Juliet Ndukum, Lu{\'\i}s~L Fonseca, Helena Santos, Eberhard~O Voit, and Susmita
  Datta.
\newblock Statistical inference methods for sparse biological time series data.
\newblock \emph{BMC systems biology}, 5\penalty0 (1):\penalty0 57, 2011.

\bibitem[Peach et~al.(2019)Peach, Yaliraki, Lefevre, and Barahona]{Peach2019}
R~Peach, S~Yaliraki, D~Lefevre, and M~Barahona.
\newblock Data-driven unsupervised clustering of online learner behaviour.
\newblock \emph{npj Science of Learning}, 4, 2019.
\newblock \doi{10.1038/s41539-019-0054-0}.
\newblock URL \url{http://dx.doi.org/10.1038/s41539-019-0054-0}.

\bibitem[Perotte and Hripcsak(2013)]{perotte2013temporal}
Adler Perotte and George Hripcsak.
\newblock Temporal properties of diagnosis code time series in aggregate.
\newblock \emph{IEEE journal of biomedical and health informatics}, 17\penalty0
  (2):\penalty0 477--483, 2013.

\bibitem[Rasmussen and Williams(2006)]{rasmussen2006gaussian}
Carl~Edward Rasmussen and Christopher~KI Williams.
\newblock \emph{Gaussian processes for machine learning}, volume~1.
\newblock MIT press Cambridge, 2006.

\bibitem[Resnik(1999)]{resnik1999semantic}
Philip Resnik.
\newblock Semantic similarity in a taxonomy: An information-based measure and
  its application to problems of ambiguity in natural language.
\newblock \emph{Journal of Artificial Intelligence Research}, 11:\penalty0
  95--130, 1999.

\bibitem[Rokach and Maimon(2005)]{rokach2005clustering}
Lior Rokach and Oded Maimon.
\newblock Clustering methods.
\newblock In \emph{Data mining and knowledge discovery handbook}, pages
  321--352. Springer, 2005.

\bibitem[Smith(2012)]{smith2012future}
Stephen~M Smith.
\newblock The future of {FMRI} connectivity.
\newblock \emph{Neuroimage}, 62\penalty0 (2):\penalty0 1257--1266, 2012.

\bibitem[Son and Baek(2008)]{son2008modified}
Young~Sook Son and Jangsun Baek.
\newblock A modified correlation coefficient based similarity measure for
  clustering time-course gene expression data.
\newblock \emph{Pattern Recognition Letters}, 29\penalty0 (3):\penalty0
  232--242, 2008.

\bibitem[Stegle et~al.(2010)Stegle, Denby, Cooke, Wild, Ghahramani, and
  Borgwardt]{stegle2010robust}
Oliver Stegle, Katherine~J Denby, Emma~J Cooke, David~L Wild, Zoubin
  Ghahramani, and Karsten~M Borgwardt.
\newblock A robust {Bayesian} two-sample test for detecting intervals of
  differential gene expression in microarray time series.
\newblock \emph{Journal of Computational Biology}, 17\penalty0 (3):\penalty0
  355--367, 2010.

\bibitem[Stewart(1998)]{stewart1998matrix}
G.W. Stewart.
\newblock \emph{Matrix Algorithms: Volume 1: Basic Decompositions}.
\newblock Society for Industrial and Applied Mathematics, 1998.
\newblock ISBN 9780898714142.
\newblock \doi{10.1137/1.9781611971408}.

\bibitem[Vinh et~al.(2010)Vinh, Epps, and Bailey]{vinh2010information}
Nguyen~Xuan Vinh, Julien Epps, and James Bailey.
\newblock Information theoretic measures for clusterings comparison: Variants,
  properties, normalization and correction for chance.
\newblock \emph{The Journal of Machine Learning Research}, 11:\penalty0
  2837--2854, 2010.

\bibitem[Yao et~al.(2005)Yao, M{\"u}ller, and Wang]{yao2005functional}
Fang Yao, Hans-Georg M{\"u}ller, and Jane-Ling Wang.
\newblock Functional data analysis for sparse longitudinal data.
\newblock \emph{Journal of the American Statistical Association}, 100\penalty0
  (470):\penalty0 577--590, 2005.

\bibitem[Yu et~al.(2010)Yu, Li, Qin, Bo, Wu, and Wang]{yu2010gosemsim}
Guangchuang Yu, Fei Li, Yide Qin, Xiaochen Bo, Yibo Wu, and Shengqi Wang.
\newblock {GOSemSim}: an {R} package for measuring semantic similarity among
  {GO} terms and gene products.
\newblock \emph{Bioinformatics}, 26\penalty0 (7):\penalty0 976--978, 2010.

\bibitem[Yu and Shi(2003)]{yu2003multiclass}
Stella~X. Yu and Jianbo Shi.
\newblock Multiclass spectral clustering.
\newblock In \emph{Proceedings Ninth IEEE International Conference on Computer
  Vision}, pages 313--319, Oct 2003.
\newblock \doi{10.1109/ICCV.2003.1238361}.

\end{thebibliography}
\end{document}